%% file: KDD20.tex
\documentclass[sigconf]{acmart}

\usepackage{dsfont}
\usepackage{bbm}

\usepackage{booktabs} 
\usepackage{amsmath}
\usepackage{subfloat}
\usepackage{caption}
\usepackage{mathtools}
\usepackage{algorithm}
\usepackage{sidecap}
\usepackage{algpseudocode}
\usepackage{multirow}
\usepackage{graphics,epsfig,float}
\usepackage{subfig,balance,lineno}
\usepackage{float}
\usepackage{lineno,multirow}
\usepackage{epstopdf}
\usepackage{footmisc}
\usepackage{url}
\usepackage{mathtools}
\usepackage{times}
\usepackage{algorithm}
\usepackage{algorithmicx}
\usepackage{algpseudocode}
\usepackage{amsmath,amsthm,amssymb}
\usepackage{mathtools}
\usepackage{dsfont}
\usepackage{algpseudocode}
\usepackage{multirow}
\usepackage{threeparttable}
\usepackage{bbm}
\usepackage{tikz}           
\usepackage{xcolor}         
\usepackage{bm}

\newlength\myindent
\newcommand{\multiline}[1]{%
  \begin{tabularx}{\dimexpr\linewidth-\ALG@thistlm}[t]{@{}X@{}}
    #1
  \end{tabularx}
}

\DeclarePairedDelimiterX{\infdivx}[2]{(}{)}{%
  #1\;\delimsize|\delimsize|\;#2%
}

\newcommand{\kld}[2]{\ensuremath{D_{KL}\infdivx{#1}{#2}}}
\newtheorem*{definition}{Definition} 
\def\BibTeX{{\rm B\kern-.05em{\sc i\kern-.025em b}\kern-.08emT\kern-.1667em\lower.7ex\hbox{E}\kern-.125emX}}

\copyrightyear{2020}
\acmYear{2020}
\setcopyright{acmcopyright}\acmConference[KDD '20]{Proceedings of the 26th ACM SIGKDD Conference on Knowledge Discovery and Data Mining}{August 23--27, 2020}{Virtual Event, CA, USA}
\acmBooktitle{Proceedings of the 26th ACM SIGKDD Conference on Knowledge Discovery and Data Mining (KDD '20), August 23--27, 2020, Virtual Event, CA, USA}
\acmPrice{15.00}
\acmDOI{10.1145/3394486.3403303}
\acmISBN{978-1-4503-7998-4/20/08}
\begin{document}
\fancyhead{}

%
\title{Automatic Validation of Textual Attribute Values in E-commerce Catalog by Learning with Limited Labeled Data  }

\author{Yaqing Wang$^{1*}$, Yifan Ethan Xu$^2$,  Xian Li$^2$, Xin Luna Dong$^2$ and Jing Gao$^1$}
\affiliation{\institution{\textsuperscript{1}State University of New York at Buffalo, Buffalo, New York \\
\textsuperscript{2}Amazon.com\\
\textsuperscript{1}\{yaqingwa,  jing\}@buffalo.edu, \\
\textsuperscript{2}\{xuyifa,  xianlee, lunadong\}@amazon.com
}
}
\thanks{$^*$Most of the work was conducted when the author was interning at
Amazon.}


%
\begin{abstract}
Product catalogs are valuable resources for eCommerce website. In the catalog, a product is associated with multiple attributes whose values are short texts, such as product name, brand, functionality and flavor. Usually individual retailers self-report these key values, and thus the catalog information unavoidably contains noisy facts. It is very important to validate the correctness of these values in order to improve shopper experiences and enable more effective product recommendation. Due to the huge volume of products, an effective automatic validation approach is needed. In this paper, we propose to develop an automatic validation approach that verifies the correctness of textual attribute values for products. This can be formulated as a task as cross-checking a textual attribute value against product profile, which is a short textual description of the product on eCommerce website. Although existing deep neural network models have shown success in conducting cross-checking between two pieces of texts, their success has to be dependent upon a large set of quality labeled data, which are hard to obtain in this validation task: products span a variety of categories. Due to the category difference, annotation has to be done on all the categories, which is impossible to achieve in real practice.  

To address the aforementioned challenges, we propose a novel meta-learning latent variable approach, called \textit{MetaBridge}, which can learn transferable knowledge from a subset of categories with limited labeled data and capture the uncertainty of never-seen categories with unlabeled data. More specifically, we make the following contributions. (1) We formalize the problem of validating the textual attribute values of products from a variety of categories as a natural language inference task in the few-shot learning setting, and propose a meta-learning latent variable model to jointly process the signals obtained from product profiles and textual attribute values. (2) We propose to integrate meta learning and latent variable in a unified model to effectively capture the uncertainty of various categories. With this model, annotation costs can be significantly reduced as we make best use of labeled data from limited categories.  (3) We propose a novel objective function based on latent variable model in the few-shot learning setting, which ensures distribution consistency between unlabeled and labeled data and prevents overfitting  by sampling different records from the learned distribution. Extensive experiments on real eCommerce datasets from hundreds of categories demonstrate the effectiveness of \textit{MetaBridge} on textual attribute validation and its outstanding performance compared with state-of-the-art approaches. 
\end{abstract}

%
%

%


\maketitle

\input{./subfiles/1_intro}
\input{./subfiles/2_methodology}
\input{./subfiles/3_experiments}

\input{./subfiles/4_related_work}

\input{./subfiles/5_conclusion}

\section*{Acknowledgment}
The authors would like to thank the anonymous referees for their valuable comments and helpful suggestions,  would like to thank Ron Benson, Christos Faloutsos, Jun Ma, Giannis Karamanolakis, Haw-Shiuan Chang, Junheng Hao, Zhengyang Wang, Yuning Mao and Wei Hao for their insightful comments on the project, and  Saurabh Deshpande,
Prashant Shiralkar, Tong Zhao for their constructive feedback on data integration for the experiments. This work is supported in part by the US National Science Foundation under grant NSF-IIS 1747614. Any opinions, findings, and conclusions or recommendations expressed in this material are those of the author(s) and do not necessarily reflect the views of the National Science Foundation.

\bibliographystyle{ACM-Reference-Format}
\bibliography{acmart}
\newpage
\input{subfiles/appendix}

\end{document}

%% file: subfiles/1_intro.tex
\section{Introduction}

Product catalogs are valuable resources for eCommerce website for the organization, standardization and publishing of product information.  
Because the majority of product catalogs on eCommerce websites (e.g., Amazon, Ebay, and Walmart) are contributed by individual retailers, the catalog information unavoidably contains noisy facts~\cite{opentag, karamanolakis2020txtract}. The existence  of  such errors results in misleading information delivered to consumers and significantly downgrades the performance of  downstream  applications, such as product recommendation. As the magnitude of product catalogs does not allow for manual validation, there is an urgent need for the development of automatic yet effective validation algorithms. 

In a product catalog, a product is typically associated with multiple textual attributes, such as name, brand, functionality and flavor, whose values are short texts. Therefore, in this paper, we focus on the important task of validating the correctness of a textual attribute value given a product. A real example is ``Ben \& Jerry's - Vermont's Finest Ice Cream, Non-GMO - Fairtrade - Cage-Free Eggs - Caring Dairy - Responsibly Sourced Packaging, Americone Dream, Pint (8 Count)'', which is the product title of an icecream on Amazon. The attribute ``flavor'' is a textual attribute, and for this particular icecream, ``Americone Dream'' is its flavor attribute value. The objective is to automatically output whether this value is correct or not for this product. 


One may consider to model this task as anomaly detection based on the values of the target textual attribute, so that anomalies correspond to wrong values.  However, this solution is not applicable to the validation task because: 1) As individual retailers self-report these attribute values, the set of possible values cannot be predetermined, and thus traditional anomaly detection approaches cannot work. 2) Textual anomaly detection has been studied and many methods have been proposed to identify anomalies by extracting distinguishing features from the texts. However, in the validation task, the correctness of a value is highly dependent on the product. For example, ``Americone dream" may not be a common piece of textual value, but it is a correct flavor name for Ben\&Jerry icecream. 

Motivated by this observation, we propose to verify the correctness of textual attribute value against the text description of the corresponding product. A detailed description of a product can be found from the product webpage, which contains rich information about many attributes of the product. For example, in our example,  the title itself already covers the values of several attributes, such as flavor and ingredients. By cross-checking the textual attribute value ``Americone-dream" for flavor against this description, we can easily verify that this value is correct. However, this cross-checking cannot be completed by a simple matching of the keywords. We found that a certain amount of errors are because the retailers often abuse the attribute by filling a real value of another attribute. Such errors cannot be detected by simply matching the value with product description text as they indeed can be found there. For example, for value ``Non-GMO", it is a wrong value as of flavor, but could be labeled as correct by a simple matching against the product title of this icecream. 

Therefore, we propose to model the validation problem as the task of automatic correctness inference based on an input of a textual attribute value and the description of the corresponding product. This setting is related to the natural language inference (NLI) task, which automatically determines if a hypothesis is true or false based on a text statement. Recently, powerful neural network based models, such as Transformer \cite{transformer} and BERT \cite{bert} have shown promising performance towards NLI task.  However, their success relies on sufficient high-quality labeled data, which requires the annotation of correctness on a large number of hypothesis-statement pairs.  This requirement cannot be satisfied in the textual attribute validation task. There are thousands to millions product categories on eCommerce website, and thus annotating sufficient labeled data for all the categories is impossible. If only limited categories are annotated, such labeled data cannot be applied to other categories. For products in different categories, the product attributes and the vocabularies of the attributes could vary significantly. For example, even for the same attribute ``flavor'', there is no overlapping values when describing the flavor of \textit{seasoning}, \textit{ice cream} and \textit{coffee}.  

To tackle the aforementioned challenges, we propose a novel meta-learning latent variable approach, namely MetaBridge, for textual attribute validation. The proposed approach effectively leverages a small set of labeled data in limited categories for training category-agnostic models, and utilizes unlabeled data to capture category-specific information. More specifically,  the proposed objective function is directly derived from the textual attribute validation task based evidence lower bound, and it seamlessly integrates meta-learning principle and latent variable modeling. We then propose to solve this problem via a stochastic neural network which has the sampling and parameter adaptation steps. The benefits of the proposed approach include the following. First, the parameter adaptation step allows more parameter flexibility to capture category-specific information. Second, we enforce the  distribution consistences between unlabeled and labeled data via KL Divergence, which makes best use of limited labeled information while extracts most useful information from unlabeled data. Third, the proposed model is a stochastic neural network where sampling step is beneficial to the prevention of overfitting. The insights behind our objective function are explored in our experiments. Experimental results on two large real-world datasets show that proposed model can effectively generalize to new product categories  and outperforms the state-of-the-art approaches.

The main contributions of this paper can be summarized as follows:
\begin{itemize}
\item We formally define the important problem of textual attribute validation on eCommerce website as an automatic correctness inference task based on a model taking an input pair of attribute-value and corresponding product description.  We propose an effective meta-learning latent variable model which can make category-specific decision even though labeled data are only collected from limited categories.

\item The proposed MetaBridge method combines meta learning and latent variable in a joint model to make best use of limited labeled data and vast amounts of unlabeled data. The proposed solution enhances the ability of capturing category uncertainty and preventing overfitting via effective sampling.   

\item We empirically show that the proposed method MetaBridge can effectively infer the correctness of attribute values and significantly outperform the state-of-the-art models on two real-world datasets collected from Amazon.com.
\end{itemize} 

The rest of the paper is organized as follows: problem setting and preliminaries are  introduced in Section~\ref{section:background},  and the details of the proposed framework are presented in Section~\ref{section:methodology}. Experimental results are presented in Section~\ref{section:experiments}. Related literature survey is summarized in Section~\ref{section:related_work}, and the study is concluded in Section~\ref{section:conclusion}.


%% file: subfiles/2_methodology.tex
\vspace{-0.05in}
\section{Problem Setting and Preliminary}
\label{section:background}

In this section, we first introduce our problem and the few-shot learning setting, then we present the representative algorithm of meta-learning, its limitations and our intuitions.


\subsection{Problem Setting}
Given a set of product profiles presented as unstructured text data like titles and their corresponding textual attribute values, our objective is to identify incorrect attribute values based on corresponding product profiles. Note that we have open world assumption thus we cannot construct a golden list to filter out never-seen attribute values.  As the the categories of product are from thousands to millions and annotation job requires corresponding knowledge, we can only obtain a small set of annotated data about a subset of product categories. But for each category,  unlabeled data are easily collected. We next formally define the problem we are solving.

\begin{definition}
 Given a set of product categories $C$ and corresponding products $I  = \{I_c: c \in C\}$,  product profiles $P = \{p_i: i \in I\}$, attribute values as $V = \{v_i: i \in I\}$, we aim to identify  $X = (P, V)$  pair that are incorrect for product $I$.
 \end{definition}
 
After defining our problem, we introduce our learning setting. Following the \textit{few-shot learning} setting~\cite{vinyals2016matching}, in each category $c \sim C$, we have a few unlabeled examples $x^s_c = \{x^s_{c,i}\}_{i=1}^{N}$ to constitute the support set $\mathcal{D}_c^{s}$ and  have a small set of labeled examples $\{x^q_c, y^q_c\}  = \{x^q_{c,i}, y^q_{c,i}\}_{i=N+1}^{N+K}$ as the query set $\mathcal{D}^{q}_c$. We need to learn from a subset of categories a well-generalized model which can facilitate training in a new category based on unlabeled support set $\mathcal{D}^{s}_c$ and infer the correctness of attribute values for corresponding products $I_c$ in the same category $c$.

\subsection{MAML} 
\label{subsection:maml}

We give an overview of Model-Agnostic Meta-Learning method~\cite{maml} which is a representative algorithm of optimization-based meta-learning approaches. First, we use our problem as an example to introduce the general learning setting of meta-learning methods. The learning of meta-learning are split into two stages: meta-training and meta-testing. During the meta-training stage, the baseline learner $f_{\theta}$ with parameter set $\theta$ will be adapted to specific category $c$ as $f_{\theta_c}$ with the help of meta-learner $M(\cdot)$ on support set $\mathcal{D}^{s}_c$, i.e., $\theta_c = M(\theta, \mathcal{D}^{s}_c)$.  Such category specific learner $f_{\theta_c}$ is evaluated on the corresponding query set $\mathcal{D}^{q}_c$. During the meta-testing stage, the baseline learner $f_{\theta}$ will be adapted to testing category $c$ on $\mathcal{D}^{s}_c$ using the same procedure with meta-training stage, i.e., $\theta_c = M(\theta, \mathcal{D}^{s}_{c})$, and make predictions for the $\mathcal{D}^{q}_{c}$.

In the MAML, it updates parameter vector $\theta$  using one or more gradient descent updates on the category $c$. For example, when using one gradient update:
$$
\theta_c = M(f_{\theta}, \mathcal{D}^{s}_c) =  \theta - \beta \bigtriangledown_{\theta} \mathcal{L}(f_{\theta}, \mathcal{D}^{s}_c),
$$
where $\beta$ is inner step size and $\mathcal{D}^{s}_c$ is a support set for given category $c$.
The model parameters are trained by optimizing for the performance of $f_{\theta_c}$ with respect to $\theta$ across categories.  More concretely, the meta-objective is as follows:
$$
\min_{\theta}  \mathcal{L}(f_{\theta}) =  \sum_{c\in C} \mathcal{L}(f_{\theta - \beta \bigtriangledown_{\theta} \mathcal{L}(f_{\theta}, \mathcal{D}^{s}_c)}, \mathcal{D}^{q}_c),
$$
where $\mathcal{D}_{c}^q$ is a query set for given category $c$.

\textbf{Limitations:} MAML captures category uncertainty with the help of a few labeled data. Such mechanism brings expensive and continuous annotation costs.  Although we can change the supervised loss on support set to unsupervised loss like entropy minimization, the adaptation on unlabeled data will undoubtedly increase the difficulty of capturing category uncertainty and further degrade the performance. Moreover, meta-learning methods suffer from overfitting problem especially when only a small set of labeled data is available. 

\textbf{Key ideas of our solution:} To avoid continuous annotation cost,  we expect our model to capture the category-uncertainty via unlabeled data. Thus, how we take advantage of unlabeled data to benefit our method is  a key problem. A simple intuition is that we need to bridge unlabeled data and labeled data together to stabilize adaptation step.  To achieve such goal, we propose a new approach which can integrate  latent variable model with meta-learning framework. The latent variable model can capture the category distribution via a latent variable which can construct a connection between unlabeled and labeled data and prevents overfitting with the inherent sampling procedure.

\section{Methodology}
\begin{figure*}[hbt]
\vspace{-0.2in}
\includegraphics[width=5in]{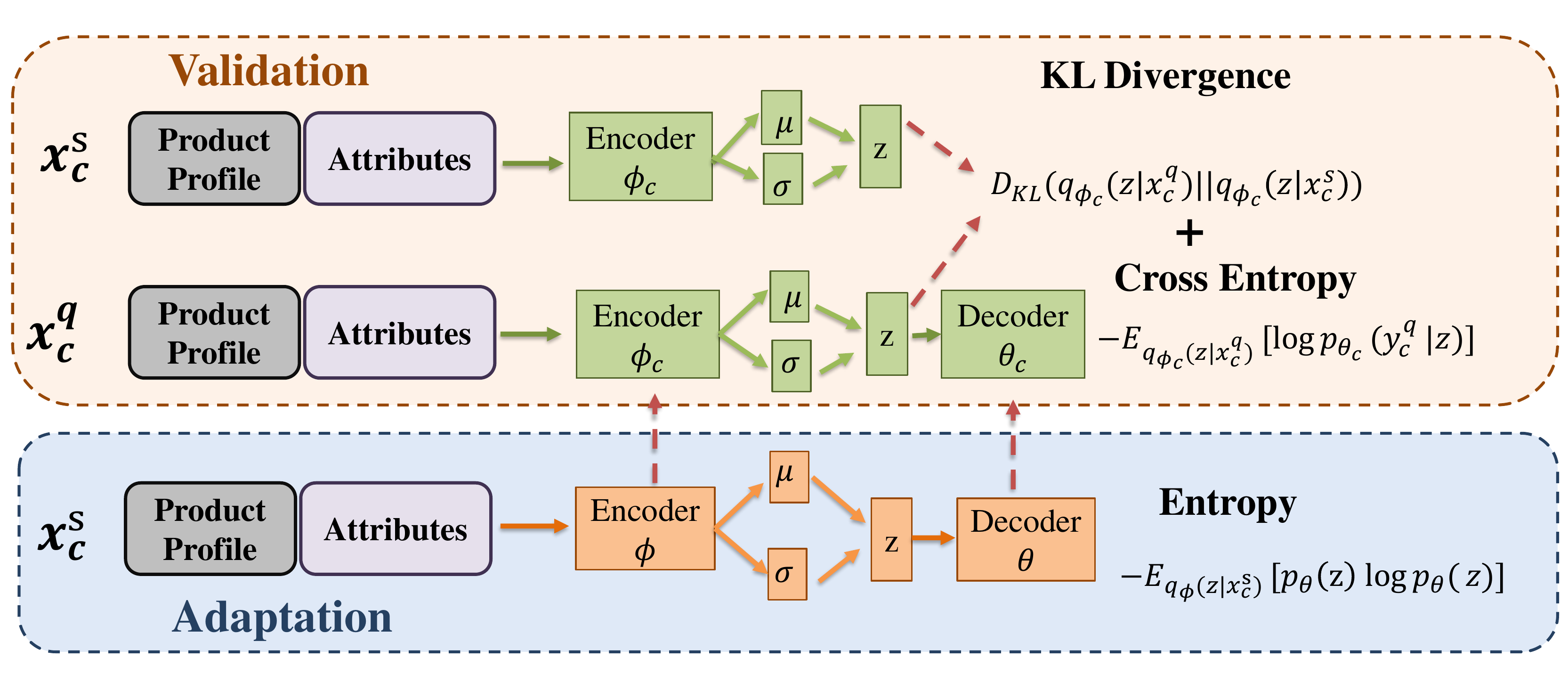}
\vspace{-0.1in}
 \caption{The proposed approach MetaBridge. The proposed approach mainly includes two stages: adaptation and Validation. During the adaptation stage, the model parameter $\Theta$ is updated to $\Theta_c$  accordingly to capture the uncertainty of category $c$. During the validation stage, the adapted model $\Theta_c$ is used to validate textual attributes for products on the category $c$.}\label{fig:framework}
\end{figure*}

\label{section:methodology}
In this section, we first introduce how we derive our meta-learning latent variable objective function, then we present our model architecture and the algorithm flow.

\subsection{Overview}

As shown in Figure~\ref{fig:framework}, the proposed MetaBridge mainly includes two stages: adaptation and validation. During the adaptation stage, the model parameter is updated on unlabeled support data from given product category; during the validation stage, the category-specific model is used to make textual validation for products from same product category.  To capture uncertainty on unlabeled data and prevent overfitting, we propose a meta learning latent variable objective function which includes two terms: inference loss and bridging regularizer. By jointly minimizing both objectives, we enforce the model i) to learn direct signal from labeled data, and ii) internally harmonizes the latent structures of the new category and existing category from unlabeled data.
More specifically, the proposed approach is a stochastic neural network which includes sampling and parameter adaptation steps. Furthermore, the proposed model can enforce the distribution consistency between unlabeled and labeled data via KL Divergence. Thus, we are able to train a complicated meta learning Transformer-based model which can jointly processes signals from textual product description and attribute values to conduct effective inference. 

\subsection{Latent Variable Model}
The goal of the proposed algorithm is to learn to infer on various categories even unseen category with a handful unlabeled training instances. More specifically, for the $c$-th category, the corresponding support set $x_c^s$ is given, we aim to infer $y_c^q$ based on $x_c^q$. Here We denote $x_c = \{x_c^s, x_c^q\}, y_c = \{y_c^q\}$ for simplicity and hence our objective function can be represented as follows:
\begin{equation} 
\begin{aligned}
\log p_{\Theta}(y|x) =  \sum_{c \in C} \log p_{\Theta}(y_c|x_c), \\
\end{aligned}
\end{equation} 
where $\Theta$ represents the parameter set of the proposed model. For each category $c$, we only have a very limited number of labeled data points. To capture category uncertainty, we include a latent variable $z$ that captures category distribution. This latent variable is of particular interest because it can capture the category uncertainty and allows us to sample data for the learned category to prevent overfitting.

To be clear, we take $c$-th category as an example. Let $p(z, y_c|x_c)$ be a joint distribution over a set of latent variables $z \sim Z$ and observed variables $y_c \in Y$ and $x_c \in X$ for category $c$. An inference query involves computing posterior beliefs after incorporating evidence into the prior: $p(z|y_c, x_c) = p(z, y_c| x_c) / p(y_c|x_c)$. This
quantity is often intractable to compute as the marginal likelihood $p(y_c|x_c) = \int_{z}p(z, y_c| x_c)dz$ requires integrating or summing over a potentially exponential number of configurations for $z$. As with variational autoencoders~\cite{vae}, we approximate the objective function using the evidence lower bound (ELBO) on the log likelihood. For the purpose of calculating ELBO,  let us introduce an encoder model $q_{\phi}(z|x_c,y_c)$: an
approximation to the intractable true posterior $p(z|x_c,y_c)$ with a parameter set ${\phi}$. 
In a similar vein, we use a decoder model $p_{\theta}(y_c|x_c,z)$  to approximate the intractable true posterior $p(y_c|x_c,z)$ with a parameter set ${\theta}$. Thus, the parameter set $\Theta$  includes $\{\phi, \theta\}$.  After introducing the encoder and decoder, we  present how to derive our objective function based on ELBO. \\

\noindent \textbf{Evidence Lower Bound (ELBO)} The ELBO can be shown to decompose into
\begin{equation} 
\begin{aligned}
\label{eq:elbo}
&\log p_{\Theta}(y_c|x_c)\\
\geq &\mathbb{E}_{q_{\phi}(z|x_c,y_c)} [ \log p_{\theta}(y_c|z, x_c)]- \kld{q_{\phi}(z|x_c, y_c)}{p(z)}.\\
\end{aligned}
\end{equation} 

To better reflect the desired
model behavior at test time, i.e., we have a handful training instances as a
support set $x_c^s$ for each category,  we explicitly split $x_c$ into support and query sets. Our goal is to model the conditional of the query set given the support set. Thus, instead of using prior $p(z)$ in Eq.~\ref{eq:elbo}, we propose to use a more informative conditional prior distribution $p(z|x_c^{s})$ as with~\cite{neuralprocess} and further rewrite our objective function as follows:
\begin{equation} 
\begin{aligned}
&\log p_{\Theta}(y_c|x_c)\\
= &\log p_{\Theta}(y_c^{q}|x_c^{s}, x_c^{q})\\
\geq & \mathbb{E}_{q_{\phi}(z|x_c^{s},x_c^{q},y_c^{q})}[\log p_{\theta}(y_c^{q}|z, x_c^{s}, x_c^{q})] \\ &- \kld{q_{\phi}(z|x_c^{q}, x_c^{s},y_c^{q})}{p(z|x_c^{s})}\\
\end{aligned}
\end{equation} 

For the encoder ${q_{\phi}(z|x_c^{s},x_c^{q},y_c^{q})}$, since $x_c^{q}$ is given and $y_c^{q}$ is implicitly encoded into parameter set $\phi$,  we assume z is conditional independent with $y_c^{q}$ given $x_c^{q}$ and $\phi$.  Thus, our objective function can be simplified as follows:
\begin{equation} 
\begin{aligned}
&\log p_{\Theta}(y_c^{q}|x_c^{s}, x_c^{q})\\
\geq &\mathbb{E}_{q_{\phi}(z|x_c^{s},x_c^{q})} [ \log p_{\theta}(y_c^{q}|z, x_c^{s}, x_c^{q})] \\ 
&- \kld{q_{\phi}(z| x_c^{s}, x_c^{q})}{p(z|x_c^{s})}
\label{eq:middle_objective_function}
\end{aligned}
\end{equation} 
 The support set $x_c^s$ is used to help the proposed model to quickly adapt to new category. Thus, how we take advantage of this set to benefit our framework is a key problem. To tackle this problem, we propose to encode the information from support set into our parameter inspired by MAML~\cite{maml} and further we can obtain a category-specific model to accelerate unseen category adaptation. We will introduce how to incorporate information from support set into our framework via parameter adaptation in the next subsection.

\subsection{Parameter Adaptation}

 As introduced in the subsection ~\ref{subsection:maml}, MAML obtains a category specific parameter set using one or more gradient descent updates based on loss from support set $x^{s}_c$. Considering the support set in our problem is unlabeled, we redefine the loss function on unlabeled support set by entropy minimization. Entropy minimization encourages the confidence of predictions and  is commonly used in the semi-supervised learning~\cite{grandvalet2005semi, lee2013pseudo,berthelot2019mixmatch} and domain adaptation~\cite{morerio2017minimal, jiang2018towards, vu2019advent}.  More concretely, the loss function $\mathcal{L}^c_s$ on the support set $x^{s}_c$ is defined by entropy as follows:
\begin{equation}
\label{eq:inner_loss}
\mathcal{L}^c_s (\theta, \phi, x^{s}_c)=  - \mathbb{E}_{q_{\phi}(z|x_{c}^{s})} [ p_{\theta}(z)\log p_{\theta}(z)]
\end{equation}
and the parameter adaptation step via one step of gradient descent is defined accordingly as follows:
\begin{equation}
\label{eq:inner_update}
\{\theta_c, \phi_c\} =  \{\theta, \phi\} - \beta \bigtriangledown_{\theta, \phi} \mathcal{L}^c_s({\theta, \phi}, x^{s}_c).
\end{equation}

Here we assume the information of support set is encoded into parameter via gradient descent and then exclude the $x^s_c$ from conditionals. Moreover, for the decoder $p_{\theta}(y_c^{q}|z, x_c^{s}, x_c^{q})$,  $y_c^q$ is conditional independent with $x_c^q$ given $z$ since $z$ is the feature representation of $x_c^q$. Thus, we can have simpler equations as follows:

\begin{equation} 
\textbf{Encoder:} \quad q_{\phi}(z|x_c^{s}, x_c^{q}) \to q_{\phi_c}(z|x_c^{q})
\end{equation}
\begin{equation} 
\textbf{Decoder:} \quad p_{\theta}(y_c^{q}|z,  x_c^{s}, x_c^{q}) \to p_{\theta_c}(y_c^{q}|z)
\end{equation}

\subsection{Objective Function}
\label{section:ojective_function}
To optimize our objective function, we still need to approximate conditional prior $p_{\theta}(z|x_c^{s})$ which is intractable. As the parameter adaptation step can encode support set into the model and captures category specific information, hence we propose to use $q_{\phi_c}(z|x_c^{s})$ as a approximation to $p(z|x_c^{s})$ and then we have our final objective function as follows:

\begin{equation} 
\begin{aligned}
&\log p_{\Theta}(y_c^{q}|x_c^{s}, x_c^{q})\\
\geq &\mathbb{E}_{q_{\phi_c}(z|x_c^{q})} [ \log p_{\theta_c}(y_c^{q}|z)] \\ 
&- \kld{q_{\phi_c}(z|x_c^{q})}{p(z|x_c^{s})}\\
\simeq &\mathbb{E}_{q_{\phi_c}(z|x_c^{q})} [ \log p_{\theta_c}(y_c^{q}|z)] \\ 
&- \kld{q_{\phi_c}(z|x_c^{q})}{q_{\phi_c}(z|x_c^{s})}
\end{aligned}
\end{equation} 

The objective function includes two terms: the first term is our supervised inference loss on query samples and the second term is to enforce conditional category distribution $q_{\phi_c}(z|x_c^q)$ consistent with conditional distribution $q_{\phi_c}(z|x_c^s)$, i.e., distributions of unlabeled and labeled data from same category.  The second term can be treated as a explicit bridge between support set and query set. $\lambda$ is a hyper-parameter that needs to be set. We explore the impact of $\lambda$ in the experiment section~\ref{subsection:hyper}.

\begin{equation}
\small
\begin{aligned}
\label{eq:final_objective_function}
  \mathcal{L}^c_q = \underbrace{-\mathbb{E}_{q_{\phi_c}(z|x_c^{q})} [ \log p_{\theta_c}(y_c^{q}|z)]}_{\textbf{Inference Loss}} + \lambda \underbrace{\kld{q_{\phi_c}(z|x_c^{q})}{q_{\phi_c}(z|x_c^{s})}}_{\textbf{Bridging Regularizer}}
  \end{aligned}
\normalsize
\end{equation}

In this paper, we assume $q_{\phi_c}(z|x_c^{q})$ and $q_{\phi_c}(z|x_c^{s})$ follow multivariate normal distributions $\mathcal{N}(\mu(x_c^{q}), \sigma^2(x_c^{q}) \mathbf{I})$ and $\mathcal{N}(\mu(x_c^{s}), \sigma^2(x_c^{s}) \mathbf{I})$ respectively. The KL Divergence $\kld{q_{\phi_c}(z|x_c^{q})}{q_{\phi}(z|x_c^{s})}$ in Eq.~\ref{eq:final_objective_function} can be analytically integrated:
\begin{equation}
\small
\label{eq:kl_divergence}
\begin{aligned}
&\kld{q_{\phi_c}(z|x_c^{q})}{q_{\phi_c}(z|x_c^{s})}\\
= &\sum_{j=1}^d \left( \log \frac{\sigma_j(x_c^{s})}{\sigma_j(x_c^{q})} + \frac{\sigma_j^2(x_c^{q})+ (\mu_j(x_c^{q}) - \mu_j(x_c^{s}))^2 }{2\sigma_j^2(x_c^{s})} -\frac{1}{2} \right),
   \end{aligned}
   \normalsize
\end{equation}
where $d$ is the dimension of $z$. Thus, we only need to calculate category loss term. To enable distribution $q_{\phi_c}(z|x_c^{q})$ differentiable, we follow previous work~\cite{vae,bengio2013estimating,bengio2014deep} to use  reparameterization trick to parameterize $z$.

\noindent \textbf{Reparameterization Trick}
 Instead of directly sampling from a complex distribution, we can reparametrize the random variable as a deterministic transformation of an auxiliary noise variable $\epsilon$. In our case, to sample from $q_{\phi_c}(z|x_c^{q})$, since $q_{\phi_c}(z|x_c^{q}) = \mathcal{N}(\mu(x_c^{q}), \sigma^2(x_c^{q}) \mathbf{I})$, one can draw samples by computing $z = \mu(x_c^{q}) + \sigma(x_c^{q}) \odot \bm{\epsilon}$, where $\bm{\epsilon} \sim \mathcal{N}(\mathbf{0}, \mathbf{I})$ and $\odot$ signify an element-wise product. By passing in auxiliary noise, our proposed model is stochastic and if we do not pass in any auxiliary noise, then the model is deterministic. 
 
 After introducing our final objective function, we will present the detailed architecture and algorithm flow in the next subsections.

\subsection{Model Architecture}
Our model mainly includes two components: encoder and decoder. 

\textbf{Encoder}  The encoder in use is Transformer~\cite{transformer}, which is a context-aware model and has been proven powerful in textual classification. The transformer takes a sequence of word tokens as
input.  In our problem, the input includes two parts: unstructured product profiles and the corresponding product textual attribute  values. As the length of two parts are usually very different, we use two Transformers to take two parts separately to obtain fixed-dimensional features. Following~\cite{bert}, the first token of every sequence is always a special classification token ([CLS]).  Accordingly, the final hidden state
corresponding to this token is used as the aggregate sequence representation. We concatenate the two final hidden states from Transformers and then feed them into two fully connected layers with weight matrix $\mathbf{W}_{\mu}^{2d \times d}$ and $\mathbf{W}_{\sigma}^{2d \times d}$ to output mean $\mu$ and $\log(\sigma)$ as suggested in~\cite{vae}.

\textbf{Decoder} The decoder is a fully connected layer with weight matrix $\mathbf{W}_o^{d \times 2}$ to take samples from inferred normal distribution and output the probability of given attribute values being incorrect.


\subsection{Training and inference procedures}
The training procedure is summarized in Algorithm \ref{alg:training}. We first sample a batch of categories and get corresponding support set and query set for each category. Given the support set , we first  update the parameter of encoder and decoder to get category-specific parameter set $\theta_c, \phi_c$ according to Eq.~\ref{eq:inner_loss} and Eq.~\ref{eq:inner_update}.  The category-specific encoder  takes query set $x_c^q$ and support set $x_c^s$ to output the parameters for the distribution $p(z|x_c^q)$ and $p(z|x_c^s)$ respectively. Then we can calculate the Bridging Regularizer in the Eq.~\ref{eq:final_objective_function}. We then sample $z's$ from the posterior $p(z|x_{t,i}^q)$ and the category-specific decoder takes $z's$ as input to infer the correctness of attribute values. Thus, our model is stochastic during the training stage. During the testing stage, the inference procedure is similar with it in the training procedure, the only difference is that for  any  data query data $x_{c,i}^q$,  its  inferred  latent  code  is  set  to  be  the conditional mean $\mu(x_{c,i}^q)= \mathbb{E}_{q_{\phi}(z|x_{c,i}^q)}[z]$ and the category-specific decoder takes $u(x_{c,i}^q)$ as input. In other words, we use the deterministic model in the testing stage to obtain stable inference results without sampling step. 

\begin{algorithm}
\caption{\bf Training Procedure.}\label{alg:training}
\flushleft
\begin{footnotesize}
\begin{algorithmic}[1]
    \Require{Task data, learning rate $\alpha$ and inner step size $\beta$;}
    \For{epoch $l \gets 1$ to $L$}
        \State {Sample a batch of categories $C$;}
        \For{all $c \in C$}
             \State{Get support set $D_c^{s}$ and query set $D_c^{q}$}
             \State{Compute loss $\mathcal{L}^c_s$ according to Eq.~\ref{eq:inner_loss}}
              \State{ Parameter fast adaptation with gradient descent:} \State{$\{\theta_c, \phi_c\} =  \{\theta, \phi\} - \beta \bigtriangledown_{\theta, \phi} \mathcal{L}^c_s({\theta, \phi}, x^{s}_c).$}
        \EndFor
        \State{ Update   $\{\theta, \phi\}  = \{\theta, \phi\} - \alpha \sum_{c \in C} \bigtriangledown_{\{\theta, \phi\}}\mathcal{L}^c_q$}
    \EndFor
\end{algorithmic}
\end{footnotesize}
\end{algorithm}

%% file: subfiles/3_experiments.tex
\vspace{-0.05in}
\section{Experiments}
\label{section:experiments}
In this section, we introduce the dataset used in the experiments, present the compared state-of-the-art baseline models, validate the effectiveness and explore insights of the proposed approach. 
\subsection{Datasets}

To fairly evaluate the performance of the proposed approach, we use two internal Amazon datasets on attributes \textit{Flavor} and \textit{Ingredient} respectively. The products in the two datatset are from thousands of product categories across different domains. When preprocessing the datasets, we first exclude the products which do not have the attribute of interest. Then we randomly select 100 products as support set and randomly select 10 products from the rest as query set in each category. We send query set to ask Amazon Mturkers to identify the correctness of attribute values based on corresponding product profiles.  Each data point is annotated by 3 Amazon Mturkers and the final label is decided by majority voting. To evaluate the performance of attribute validation models for never-seen product categories, we split the datasets into the training, validation, testing sets according to their product categories. Thus, we ensure that they do not contain any common product category.  To evaluate the performance of models under a small data setting, we only use a small portion of product categories for training purpose and the number of product category in training, validation and testing are in a 3:1:6 ratio.  
The detailed statistics are shown in Table~\ref{tab:stat}. 

\begin{table}[htb]
\vspace{-0.1in}
\centering
  \caption{The Statistics of the Amazon Datasets.}
  \label{tab:stat}

  \resizebox{\columnwidth}{!}{\begin{tabular}{c|c|c|c}
    \toprule
     Dataset&\# of Product Categories & \# of unlabeled Data & \# of labeled Data\\
    \hline
    Flavor& 321 & 32,100 & 3,210\\
     \hline
     Ingredient& 658 & 65,800 
     &  6,580 \\
     \bottomrule
\end{tabular}}
\vspace{-0.05in}
\end{table}
\vspace{-0.1in}

\subsection{Experimental Setup}

\noindent\textbf{Metric.} 
We use  Precision-Recall AUC (PR AUC) and Recall@Precision (R@P) to evaluate the performance of the models. PR AUC is defined as the area under the precision-recall curve. Such a metric is a useful measurement  of prediction when the classes are imbalanced. R@P is defined as the recall value at a given precision. Such a measure is widely used to evaluate the model performance when a  specific precision requirement need to be satisfied.


\noindent\textbf{Baselines.}
To validate the effectiveness of the proposed model, we choose baselines from the following three categories: supervised learning, fine-tune and meta-learning settings.

$\bullet$\textbf{Supervised Learning} 
We use Logistic Regression (LR), Support Vector Machine (SVM) and Random Forest (RF) as baselines.  The supervised learning models are only trained with labeled query data and are not updated when testing. 
The feature vectors are formed by concatenation of counting the frequencies of
specific attribute value in the product textual description, the position of first appearance of attribute value in the description and the average of attribute value word embeddings.

$\bullet$\textbf{Fine-tune} Attribute validation is related to natural language inference~(NLI) problem. We select three state-of-the-art models ESIM~\cite{esim}, Transformer~\cite{transformer}, BERT~\cite{bert} as baselines. 
All sublayers of ESIM produce the output with dimension $d=16$ except the last output layer. For the BERT model, we use the output from BERT-base's last second layer and feed the output into a fully connected layer with weight matrix $\mathbf{W}^{768 \times 16}$ with ReLU activation function. The Transformer architecture is described in detail in subsection~\ref{section:implementation}. Then the output goes through a fully connected layer to output inference results. In the fine-tune setting, the training data include unlabeled support data and labeled query data. We use the
entropy minimization to define the loss on unlabeled data as
~\cite{grandvalet2005semi} and use the cross-entropy to define the loss on labeled data. The ratio of labeled loss and unlabeled loss is set as 10:1. In the testing stage, the pre-trained model is first fine-funed on the unlabeled support data of given task with entropy minimization, and then conduct inference on testing query data. 

$\bullet$\textbf{Meta-Learning} We select two state-of-the-art meta learning models  MAML~\cite{maml} and  Meta-SGD~\cite{metasgd} as baselines. The model architectures of two baselines are identical with Transformers in fine-tune setting.  The meta learning setting is that we use entropy minimization loss on unlabeled support data to adapt the parameter of models to given tasks, the task-specific parameters will be evaluated on the query data from same task during training stage.  In the testing stage, the baselines is first fine-funed on the unlabeled support data with fixed steps of gradient updates and then conduct inference on the testing query data. 
\subsection{Implementation Details}
\label{section:implementation}
The 300  dimensional FastText pre-trained word-embedding weights~\cite{fasttext} are used to initialize the parameters of the word embedding layer for  deep learning models except for BERT. The encoder of our model is based on Transformer.  For our encoder design,   we first remove the decoder of original Transformer and only keep Transformer's encoder part. Then we change 6 identical layers of original Transformer's encoder to one instead. The multi-head number is set to 2. All sub-layers of our encoder produce outputs of dimension $d=16$ and the dropout rate is selected as 0.3 based on validation set.  The identical architecture with our proposed model is employed for the baselines Transformer, MAML and Meta-SGD. The main difference between baselines and our model in architecture is that out model have sampling step and baselines are deterministic.  We implement all the deep learning baselines and the proposed approach with PyTorch 1.2. For training models, we use Adam~\cite{adam} optimizer in the default setting. The learning rate $\alpha$ is 0.0001.  We use mini-batch size of 64 and training epochs of 400.  The parameter gradient update step is set to 1 and inner learning rate $\beta$ is set to 0.3 for all fine-tune and meta-learning models. The traditional models (LR, SVM, RF) are implemented by scikit-learn package~\cite{scikit-learn}.  The best parameters are selected based on the validation set.

\vspace{-0.1in}
\subsection{Performance Comparison}
Table~\ref{tab: expermental_results} shows the performance of different approaches on the \textit{Flavor} and \textit{Ingredient} datasets. We use 100 unlabeled data as support set and 5 labeled data as query set per product category. We can observe that that the proposed framework achieves the best results in terms of all the evaluation metrics on both datasets. 

\begin{table*}[htb]
\centering
\caption{The performance comparison of different methods in the Flavor and Ingredient data. }
\vspace{-0.05in}
\label{tab: expermental_results}
\resizebox{\textwidth}{!}{
\begin{tabular}{c|c|c|c|c|c|ccc|c|c|c|c}
\toprule
\multirow{2}{*}{Setting}&\multirow{2}{*}{Method}&\multicolumn{5}{c}{Flavor} && \multicolumn{5}{c}{Ingredient}\\
\cline{3-7} \cline{9-13}
& &  PR AUC & {R@P=0.7} & {R@P=0.8} &{R@P=0.9}  & {R@P=0.95} &&PR AUC & {R@P=0.7} & {R@P=0.8} &{R@P=0.9}  & {R@P=0.95} \\
\cline{1-13}
\multirow{3}{*}{Supervised Learning} &LR & 0.6830 $\pm$ 0.0000&		48.67 $\pm$ 0.00&		23.24	 $\pm$ 0.00&0.00	 $\pm$ 0.00&	0.00 $\pm$ 0.00 & &0.4520 $\pm$ 0.0000	 &	18.71 $\pm$ 0.00&		14.08 $\pm$ 0.00&	11.67 $\pm$ 0.00&		11.47 $\pm$ 0.00\\
&SVM & 0.6408 $\pm$ 0.0000	&	42.37 $\pm$ 0.00& 	13.56 $\pm$ 0.00&	0.00 $\pm$ 0.00&	0.00 $\pm$ 0.00 && 0.3863	$\pm$ 0.0000	&		19.72 $\pm$ 0.00&		3.22 $\pm$ 0.00&		3.22 $\pm$ 0.00&		3.22 $\pm$ 0.00\\
&RF & 0.6986 $\pm$	0.0095  &	43.78 $\pm$	1.53 &	15.81 $\pm$	5.88 &	4.43 $\pm$	2.81 &	2.45 $\pm$	2.18 && 0.4683 $\pm$	0.0137	&	20.72 $\pm$	1.33 &	16.15 $\pm$	1.49 &	14.69 $\pm$	1.06 & 11.07 $\pm$	1.28\\
\cline{1-13}
\multirow{5}{*}{Fine-tune}
&RNN&0.7092 $\pm$	0.0155	& 51.09 $\pm$	5.68	& 34.14 $\pm$	2.85	& 15.93 $\pm$	4.09	& 8.35 $\pm$	2.36 &&  0.4388 $\pm$	0.0134
&25.88 $\pm$	2.29
&20.68 $\pm$	2.49
&14.69 $\pm$	2.98
&7.69 $\pm$	4.61\\
&ESIM& 0.7160 $\pm$	0.0192	&	54.90 $\pm$	5.26 &	38.32 $\pm$	5.09 &	22.22 $\pm$	5.92 &	7.69 $\pm$	6.62  && 0.4412 $\pm$	0.0199  &	23.30 $\pm$	6.42 &	16.46 $\pm$	6.95 &	8.89	$\pm$ 5.45 &	5.07 $\pm$	3.88\\
&Transformer& 0.7210 $\pm$	0.0434	&54.19 $\pm$	10.97	&34.21 $\pm$	10.27	&19.39 $\pm$	6.72	&12.86 $\pm$	3.91 && 0.4890 $\pm$	0.0203 &	31.47 $\pm$	2.46 &	28.05 $\pm$	2.94 &	22.90 $\pm$	2.94 &	11.31 $\pm$	8.77\\
&BERT &0.7599 $\pm$ 0.0054 &63.72 $\pm$ 1.27 &	45.56	 $\pm$  3.86	& 27.76 $\pm$ 	2.34 &18.52	 $\pm$ 2.76 && 0.5292 $\pm$	0.0111	& 34.00 $\pm$	1.21	& 28.17 $\pm$	1.61	& 17.00 $\pm$	3.92	& 13.08 $\pm$	6.04\\
\cline{1-13}
\multirow{3}{*}{Meta-Learning}
&MAML & 0.7486 $\pm$	0.0128	&	61.07 $\pm$	2.55 &	39.66 $\pm$	3.48 &	22.62 $\pm$	4.19 &	15.57 $\pm$	3.71 && 0.5289 $\pm$	0.0247  &	34.46 $\pm$	2.43 &	29.73 $\pm$	3.44 &	22.48 $\pm$	6.41 &	16.05 $\pm$	6.16\\
&Meta-SGD& 0.7575 $\pm$ 0.0126 & 64.19 $\pm$ 3.51 & 42.10 $\pm$ 4.62 & 25.06 $\pm$ 2.83 & 15.01 $\pm$ 4.64 && 0.5312 $\pm$	0.0141	&	32.80 $\pm$	3.43 &	24.95 $\pm$	1.18 &	22.40 $\pm$	1.19 &	20.59 $\pm$	1.34\\
\cline{2-13}
&{MetaBridge} & \textbf{0.7852 $\pm$	0.0027}	&	\textbf{69.49 $\pm$	0.99} &	\textbf{50.00 $\pm$	1.86} &	\textbf{30.77 $\pm$	1.52} &	\textbf{22.64 $\pm$	2.37}  && \textbf{0.5658 $\pm$	0.0077} & 	\textbf{39.24 $\pm$	1.60} &	\textbf{34.57 $\pm$	2.22} &	\textbf{27.00 $\pm$	0.82} &	\textbf{21.97 $\pm$	3.52} \\
\bottomrule
\end{tabular}
}
\end{table*}

On the Flavor dataset, the LR, SVM and RF achieves the similar performance compared with RNN. The results show that the traditional models can achieve comparable performance with deep learning models when a small set of labeled data is given. Among the fine-tune models, we can observe that BERT achieves the better performance compared with RNN, ESIM and Transformer. The main difference between BERT and other baselines lies in the embedding. The improvement suggests the pre-trained embedding of BERT is informative. The RNN, ESIM and Transformer use the same pre-trained fasttext word embedding layer. The comparison between the three baselines indicate that Transformer architecture can take advantage of training data effectively compared with other two baselines.  For the meta-learning setting, we can observe that MAML achieves more than 2\% improvement in terms of PR AUC compared with Transformer with identical structure. The reason is that MAML can achieve a base parameter which can easily adapt to new task compared with semi-supervised loss learning. Besides a good base parameter,  Meta-SGD also learns update directions and learning rates during training procedure. Thus, Meta-SGD achieves better performance compared with vanilla MAML. It is worth noting that the Meta-SGD achieves comparable performance with the best baseline BERT but uses much less parameters.  The proposed approach MetaBridge achieves 3.66\% improvement over Meta-SGD and 3.33\%  compared with BERT respectively in terms of PR AUC. The improvement can also be observed from recall at given precision. Since R@P=0.8 is similar with annotators' precision, we also compare the approaches in terms of this metric. The proposed framework achieves more than 10\% improvement compared with best baseline BERT in terms of R@P=0.8. 

On the Ingredient dataset,  the RF achieves better performance compared with deep learning models RNN and ESIM. This further reveals the challenges of deep learning model in the small data learning setting. Among fine-tuned models,  similar results can be observed as those in the Flavor dataset. BERT achieves the best performance compared with other fine-tuned models. This result confirms the effectiveness of pre-trained procedure in the small data learning setting. However, a contradict result with Flavor dataset can be observed from comparison between BERT and Meta-learning models. The MAML and Meta-SGD achieves the comparable and even better performance with BERT. The reason is that the vocabularies of  ingredients are rarely used in other contexts hence the information is difficult to be captured without training on the given task dataset. This improvement shows the potentials of meta-learning models for the downstream tasks, which needs models to rapidly learn with a small set of data. Accordingly, the proposed framework achieves  6.98\% improvement in terms of PR AUC compared with BERT. Compared with best baseline Meta-SGD, the proposed framework achieves 6.51\% in terms of PR AUC. The similar improvement can be also observed from performance comparison on R@P=0.8, the proposed framework improves more than 16\% compared with the second best result.  Furthermore, we can observe that the proposed MetaBridge achieves the
best performance compared with all the baselines.
\vspace{-0.1in}
\subsection{Ablation Study}

Compared with MAML, our derived objective function has two main differences: stochastic characteristic and KL Divergence between support and query data. Thus, we are interested in their roles in the performance improvements. As introduced in the Section~\ref{section:ojective_function}, we cannot simply remove one of them  considering the KL Divergence and sampling are tightly coupled with each other. Instead, we propose two variants of MAML as baselines to explore the role of stochastic and KL Divergence respectively.  To explore the role of stochastic characteristic, we add random noise into the input to last layer of MAML  and denote it as stochastic variant. To explore the role of KL Divergence, we reduce sampling step and assume that the posterior distributions of support and query data are from normal distributions with fixed variances $\mathcal{N}(\mu(x_t^{s}), 1)$ and $\mathcal{N}(\mu(x_t^{q}), 1)$. The proposed variant is denotes as KL variant. 

\begin{figure}[hbt]
\vspace{-0.1in}
\includegraphics[width=2.in]{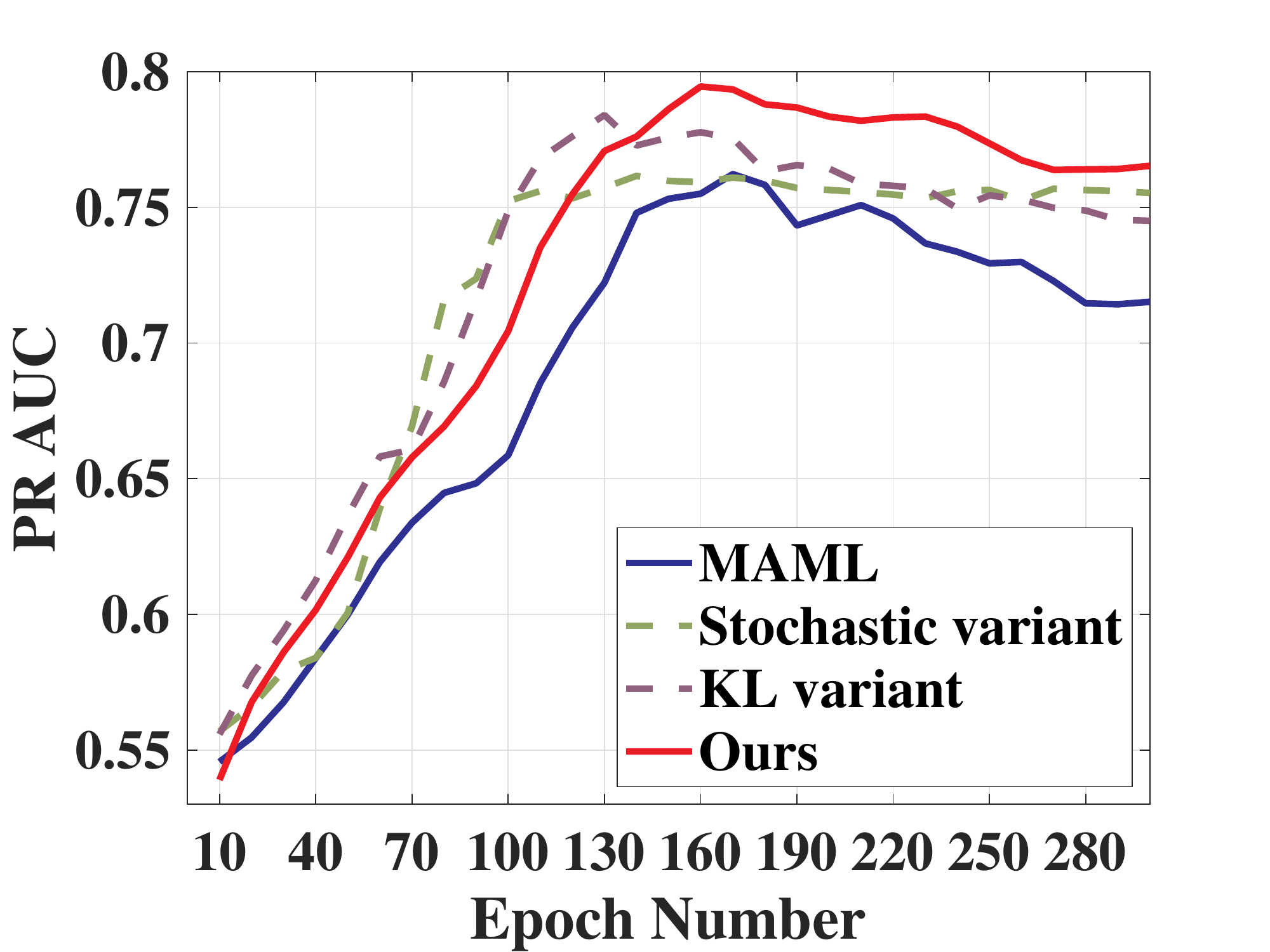}

 \caption{The changes of PR AUC for the models in term of the number of Epochs.}\label{fig:insight}
 \vspace{-0.1in}
\end{figure}

We use Flavor dataset as an example. As can be seen from Fig.~\ref{fig:insight}, the highest PR AUC score of stochastic variant is similar with that of MAML. However, unlike MAML, the stochastic variant remains highest value without dropping. This shows that the stochastic characteristic can help prevent overfitting issue. By the comparison between KL variant and MAML, we can observe the KL variant can achieve a better PR AUC during the all training epochs. This shows the KL Divergence can construct an effective information flow between support and query data to further improve the performance. However, the KL variant simply assumes that posterior distributions are from normal distribution with fixed variances, and the over-simplistic assumption limits the potential of KL Divergence. By incorporating variances estimation, our proposed framework avoids the over-simplistic distribution assumption and can achieve better performance compared with KL variant. In overall, our proposed framework enjoys the benefits of stochastic characteristic and KL Divergence simultaneously. 

\subsection{Hyperparameter Analysis}
\label{subsection:hyper}
\begin{figure}[hbt]

\label{fig:insights}
\includegraphics[width=2.in]{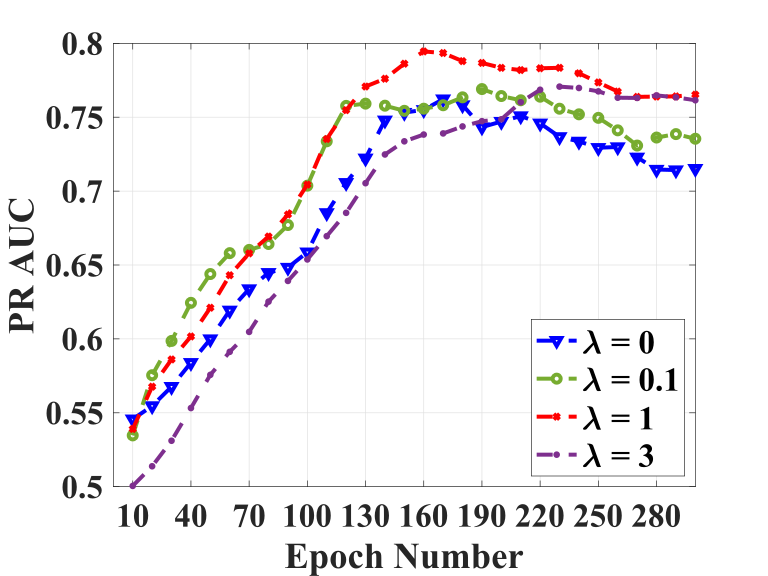}

 \caption{The changes of PR AUC with different $\lambda$'s.}\label{Fig:lambda}
\end{figure}

In our objective function, we use hyperparameter $\lambda$ to control the strength between Inference loss and KL Divergence. In this study, we aim to explore the impact of $\lambda$ in the proposed framework. We train the proposed framework using different hyperparameter $\lambda$ on the Flavor Dataset. Fig.~\ref{Fig:lambda} shows the PR AUC changes of the proposed model with respect to different $\lambda$'s. When $\lambda$ is set to 0, the sampling procedure is removed and the model is equivalent to MAML. We can observe that such a variant cannot effectively take advantage of unlabelled support data and the best PR AUC score is lower than that of other approach variants. And, such a variant suffers from the overfitting issue and converges to worst PR AUC value compared with other models. After changing the $\lambda$ from 0 to 0.1, we can observe that the PR AUC values are stably higher than that of the variant with $\lambda = 0$. As the $\lambda$ value further increases from 0.1 to 1, the proposed framework  achieves significant improvement around 4\% in terms of PR AUC compared with the variant $\lambda = 0$. This illustrates that our objective function can take advantage of unlabeled and small labeled data effectively and  improves the generalize ability of the model. When we change value of $\lambda$ to 3, the PR AUC of model increases slowly in the first 220 epochs compared with other models. But after 220 epochs, the model can archive a high PR AUC value. This further confirms the effectiveness of KL Divergence.

\subsection{Varying Size of Labels}
\begin{figure}[hbt]

\subfloat[PR AUC]{
\begin{minipage}{.23\textwidth}
\includegraphics[height=1.3in]{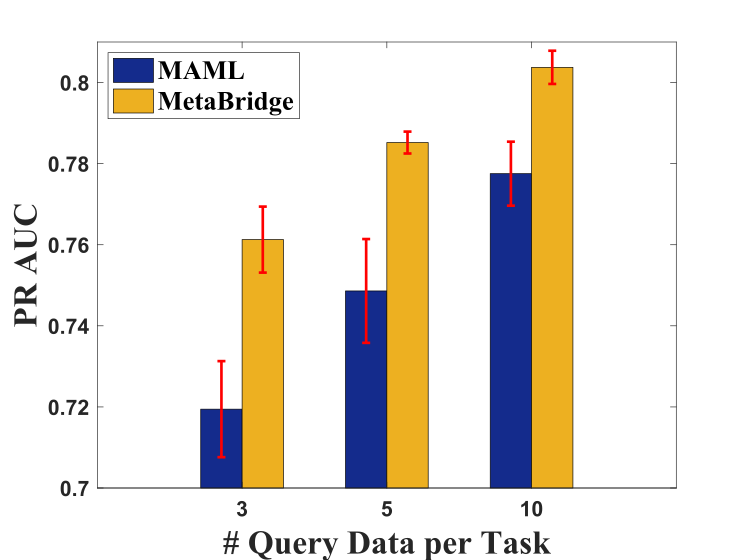}
\label{subfig:precision_score}
\end{minipage}}
\subfloat[R@P=0.8]{
\begin{minipage}{.23\textwidth}
\includegraphics[height=1.3in]{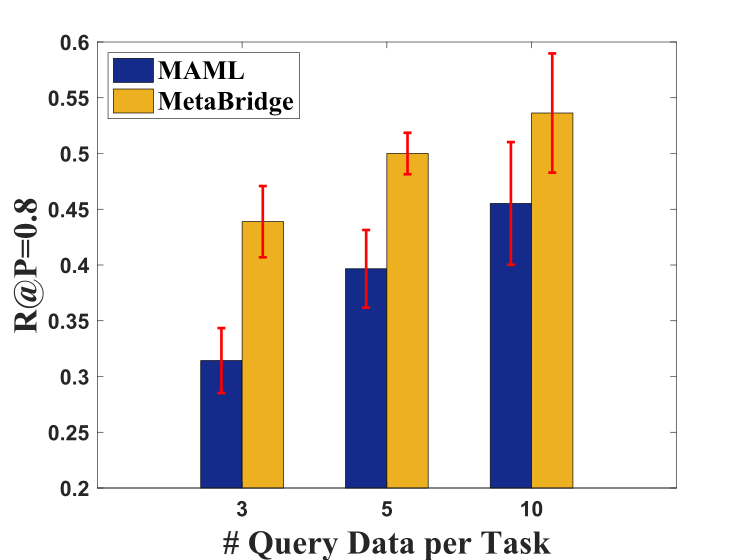}
\label{subfig:recall_precision}
\end{minipage}}
 \caption{The performance comparison of models with different numbers of query data per product category.}\label{fig:query_data_num}
\end{figure}

To analyze the impact of the query data size per product category, we train the proposed approach with different number of query data as 3, 5, 10 per category. The procedure is repeated five times and we report average performance with corresponding standard deviation. To be simple, we denote model variant by its name and number of query data. For example, the MAML which is trained with 3 query data per category is denoted as MAML$_3$.  Figure~\ref{fig:query_data_num} shows the performance comparison of the models with different number of query data in terms of PR AUC (Fig.~\ref{subfig:precision_score}) and R@P=0.8 (Fig.~\ref{subfig:recall_precision}). When  query data number is 3, our proposed framework achieves around 5.5\% improvement compared with MAML$_3$ in terms of PR AUC. This demonstrates the effectiveness of our model with a smaller set of labeled data available. The reason is that our proposed framework can caputre category uncertainty via unlabeled data  and enforce distribution consistence between unlabeled support and labeled query data. Thus, the improvement of our proposed framework over MAML is larger when the number of query data is smaller.  As the number of query data increases, the performance values of MAML and our proposed framework improve significantly. This shows that meta-learning models can effectively take advantage of labeled data. For all three settings, our proposed framework shows significant improvement compared with MAML. The improvement further confirms the superiority of our proposed framework.  

The similar results can be observed from Fig.~\ref{subfig:recall_precision}. The R@P is an important metric when we evaluate our model in the real setting. Our model achieves around 40\% and 30\% improvement respectively over MAML in terms of R@P=0.8 when the number of query data is set to 3 and 5. When the number of query data is set to 10, the R@P of our model is 53.6\%  which is higher than that of our proposed framework with 5 query data more than 6\%. This reveals the potential of our model if more labeled data is available.

%% file: subfiles/4_related_work.tex
\section{Related Work}
\label{section:related_work}

Attribute validation task is related to anomaly detection which aims to find patterns in data that do not conform to expected behavior~\cite{anomalysurvey1}.  In the anomaly detection, the most related line of research is log anomaly detection which aims to find text, which can indicate the reasons and the nature of the failure of a system~\cite{anomalysurvey2}. The traditional methods typically extract features from unstructured texts and then detect anomalies based on hand-craft features. Compared with traditional learning, deep learning  models  have  achieved  an  improvement  in the performance of anomaly detection due to their powerful abilities~\cite{anomalysurvey2}. The deep learning anomaly detection (DAD) approaches~\cite{ A_CNN2, A_RNN} model the log data as a natural language sequence and apply RNN and CNN to detect anomalies. Different with log anomaly detection, our problem needs to infer the  correctness of attribute values based on product profile information.

Attribute validation task is also related to natural language inference (NLI).  NLI is a classification task where a system is asked to classify the relationship between a pair of premise and hypothesis as either entailment, contradiction or neutral.  Large annotated datasets such as the Stanford Natural Language Inference ~\cite{nli_stanford} (SNLI) and Multi-Genre Natural Language Inference~\cite{multi_nli} (MultiNLI) corpus have promoted the development of many different neural NLI models~\cite{transformer, esim, bert, ghaeini2018dr} that achieve promising performance. However, NLI task usually requires large annotated datasets for training purpose. While pre-training is beneficial, it is
not optimized to allow fine-tuning with limited supervision and such models can still require large amounts of task-specific data for fine-tuning~\cite{bansal2019learning, yogatama2019learning}. Thus, how to train a NLI model with a small set of dataset for a specific domain is still a very challenging problem.

Another related and complementary line of research is meta-learning. Meta-learning has long been proposed as a form of learning that would allow systems to systematically build up and re-use knowledge across different but related tasks~\cite{metalearning_survey}. More specifically, meta-Learning approaches can be broadly classified into three categories: optimization-based, model-based and metric-learning based models. Optimization-based methods aim to modify the gradient descent based learning procedure for   new task quick adaptation. In the optimization-based methods, MAML~\cite{maml} is a recent promising model which learns a set of model parameters that are used to rapidly learn novel task with a small set of labeled data. Following this direction, Meta-SGD~\cite{metasgd} learns step sizes and updates directions besides initialization parameters in the training procedure. In our problem, we propose to use unlabeled data to capture task uncertainty to avoid constant labeling efforts. Towards this end, our objective function has two terms: inference loss and bridging regularizer. The proposed objective function constructs an effective information flow between unlabeled support data and labeled query data via the bridging regularizer and the sampling procedure can further prevent overfitting. Thus,  we can train a Transformer-like models to capture rich feature representations for specific task with limited labeled data and the proposed framework can quickly adapt to new task with unlabeled data only.

%% file: subfiles/5_conclusion.tex
\section{Conclusions}
\label{section:conclusion}
In this paper, we proposed to solve an important but challenging problem faced by many eCommerce websites, which is the automatic validation of textual attribute value associated with a product. Due to the large number of product categories and the huge variety among products in different categories, we cannot obtain sufficient training data for all the categories, which are needed for training deep learning models. In light of this challenge, we proposed a novel meta-learning latent variable approach, MetaBridge,  that can leverage labeled data for limited categories and utilize unlabeled data for effective correctness inference. The proposed model captures category-uncertainty via unlabeled data and trains a Transformer-based model with limited labeled data. The proposed framework has shown significantly improved performance in textual attribute validation. This was demonstrated in a series of experiments conducted on two real-world datasets from hundreds of product categories across domains on Amazon.com.


%% file: subfiles/appendix.tex
\appendix